\documentclass[letterpaper, 10 pt, conference]{ieeeconf}  

\IEEEoverridecommandlockouts                              

\usepackage{algorithm}
\usepackage{algorithmic}
\usepackage{url}
\usepackage{amsmath}
\usepackage{cite}
\usepackage[dvips]{graphicx}
\usepackage{epsfig}
\usepackage{epic,eepic,color}
\usepackage{times}
\usepackage{mathptmx}
\usepackage{multirow}
\usepackage{stkernel}

\usepackage{cases}

\usepackage{times}
\usepackage{multirow}

\definecolor{red}{rgb}{1,0,0}
\definecolor{green}{rgb}{0,1,0}
\definecolor{blue}{rgb}{0,0,1}
\definecolor{violet}{rgb}{1,0,1}
\definecolor{cyan}{cmyk}{1,0,0,0}
\definecolor{magenta}{cmyk}{0,1,0,0}
\definecolor{yellow}{cmyk}{0,0,1,0}
\definecolor{white}{rgb}{1,1,1}
\definecolor{black}{rgb}{0,0,0}

\definecolor{white}{rgb}{1,1,1}

\newcommand{\CO}[1]{}

\usepackage{jumoline}

\newcommand{\CommentOut}[1]{}

\newcommand{\FIG}[3]{
\begin{minipage}[b]{#1cm}
\begin{center}
\includegraphics[width=#1cm]{#2}
{\scriptsize #3}
\end{center}
\end{minipage}
}

\newcommand{\FIGR}[3]{
\begin{minipage}[b]{#1cm}
\begin{center}
\includegraphics[angle=-90,clip,width=#1cm]{#2}\vspace*{1mm}
\\
{\scriptsize #3}
\vspace*{1mm}
\end{center}
\end{minipage}
}

\newcommand{\FIGpng}[5]{
\begin{minipage}[b]{#1cm}
\begin{center}
\includegraphics[bb=0 0 #4 #5, clip, width=#1cm]{#2}\vspace*{-1mm}\\
{\scriptsize #3}
\vspace*{1mm}
\end{center}
\end{minipage}
}

\CO{
\renewcommand{\FIGpng}[5]{\ \ }
\renewcommand{\FIGR}[3]{\ \ }
}

\newcommand{\noeditage}[1]{#1}
\newcommand{\editage}[1]{}

\CO{
\onecolumn
\textwidth=15cm
\renewcommand{\noeditage}[1]{}
\renewcommand{\editage}[1]{#1}
}

\begin{document}

\author{Tanaka Kanji
\thanks{Our work has been supported in part by JSPS KAKENHI Grant-in-Aid for Young Scientists (B) 23700229, and for Scientific Research (C) 26330297.}
\thanks{K. Tanaka is with Graduate School of Engineering, University of Fukui, Japan.
{\tt\small tnkknj@u-fukui.ac.jp}}%
\vspace*{-5mm}}

\thispagestyle{empty}
\pagestyle{empty}

\newcommand{\figQ}{
  \begin{figure}[b]
\vspace*{-5mm}
\FIG{8}{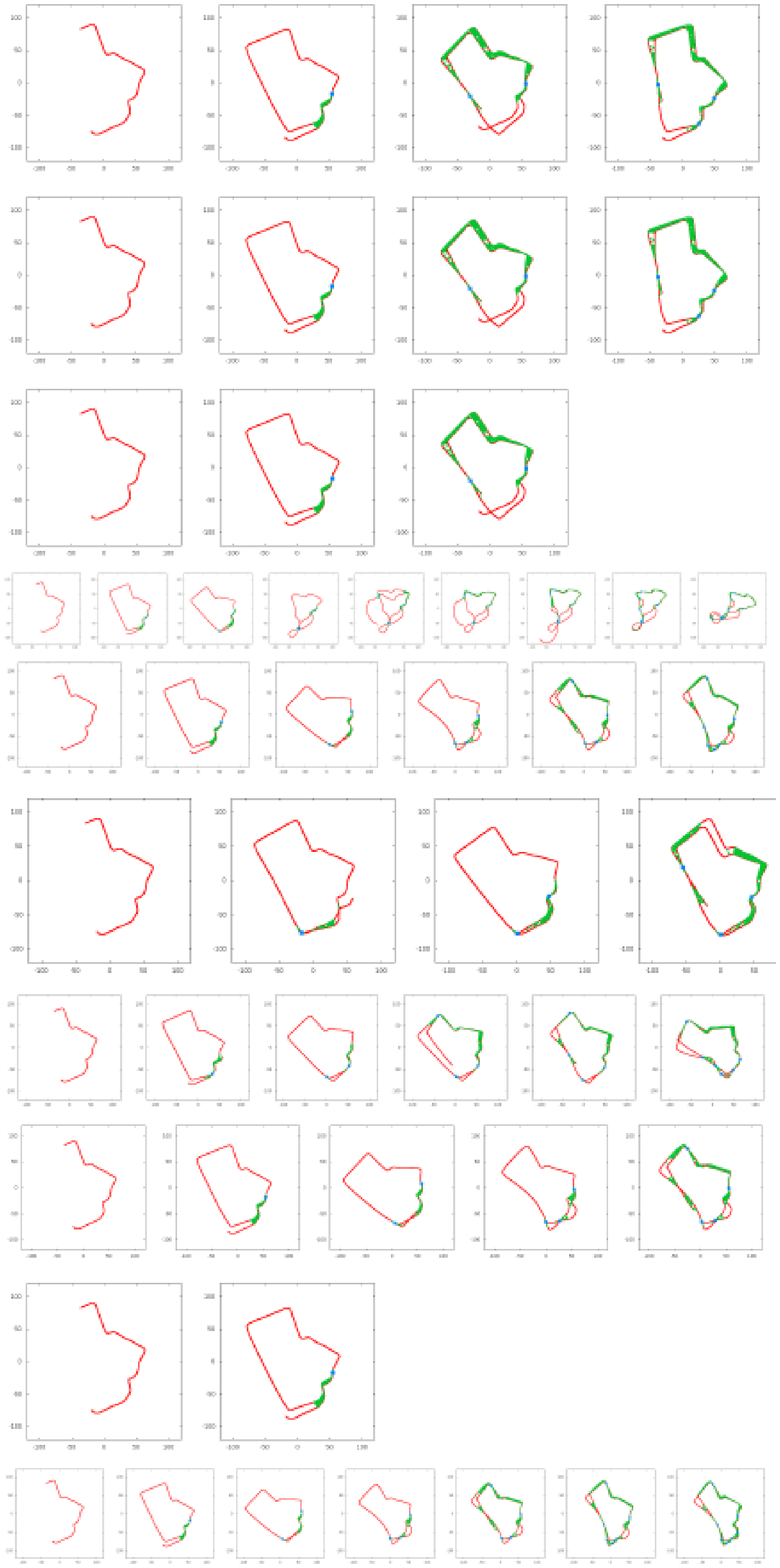}{}
\caption{Iterative process of selecting constraints.
  Each $r$-th row corresponds to the hypothesis 
that received an $r$-th rank at the end of the navigation.
  For each row,
  from left to right,
  each $i$-th panel shows
  the trajectory estimated by using $1, \cdots, i-1$ constraints selected.
}\label{fig:Q}
\end{figure}
}

\newcommand{\figR}{
  \begin{figure*}[t]
\FIG{17}{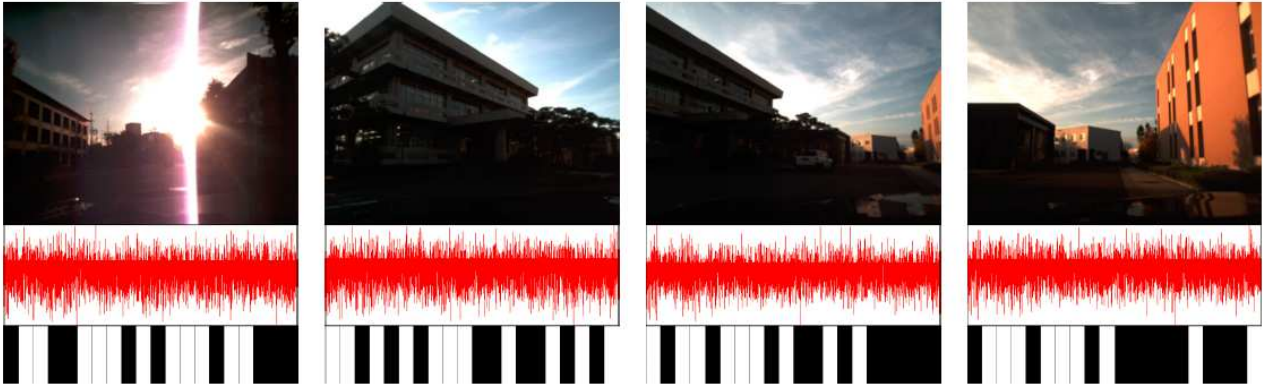}{}
\caption{Examples of visual features used by the image retrieval system.
  For each figure, the top, middle and bottom panels
  are input image, the 4096-dimensional DCNN feature and its binary code. }\label{fig:R}
\vspace*{-5mm}
\end{figure*}
}

\newcommand{\figS}{
  \begin{figure}[b]
\FIGR{8}{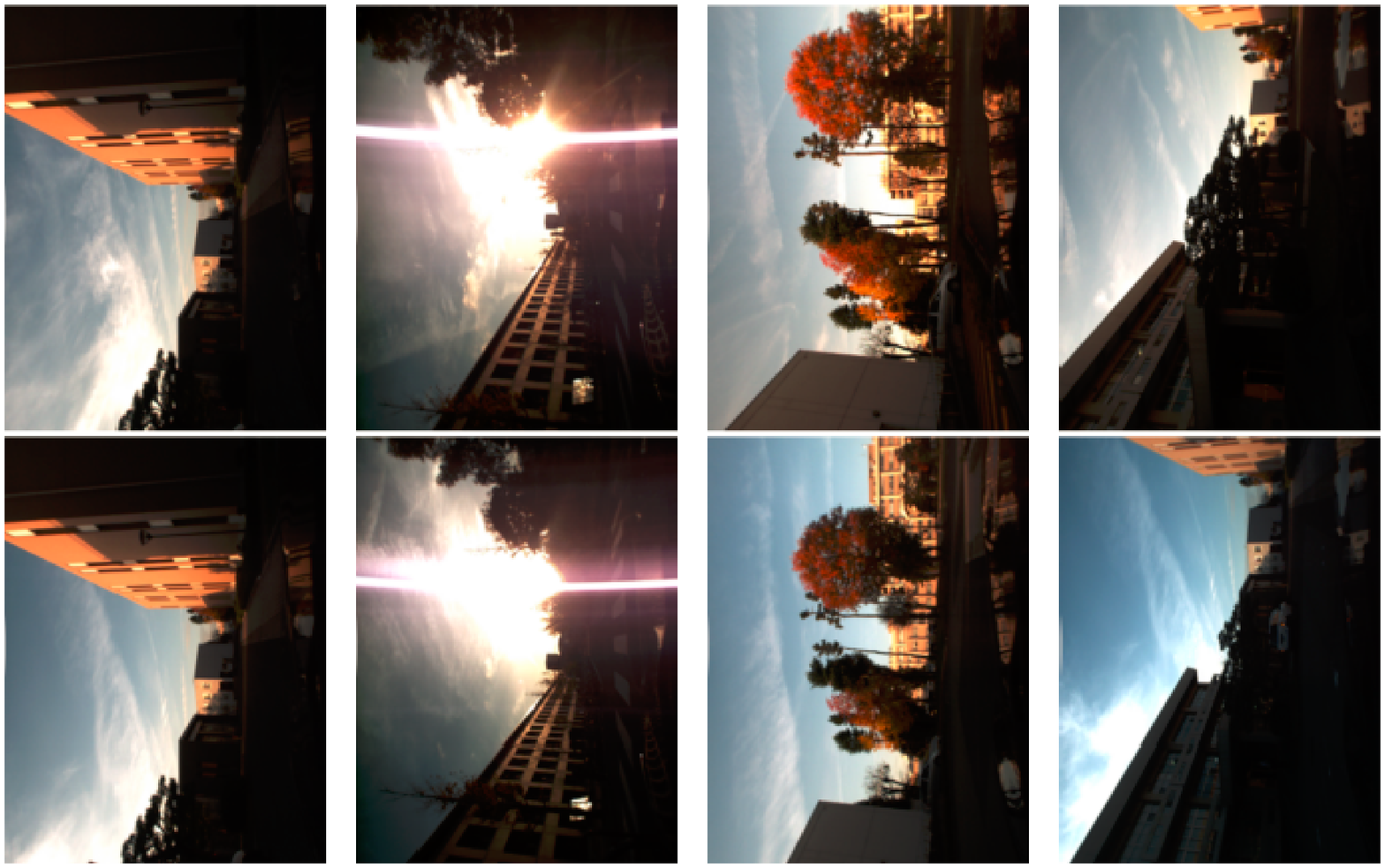}{}\vspace*{-5mm}\\
\caption{Loop closure constraints detected by
  the proposed method.
  Left: query image. Right: retrieved image.
}\label{fig:S}
\end{figure}
}

\newcommand{\figT}{
  \begin{figure}[b]
\vspace*{-5mm}
\FIG{8}{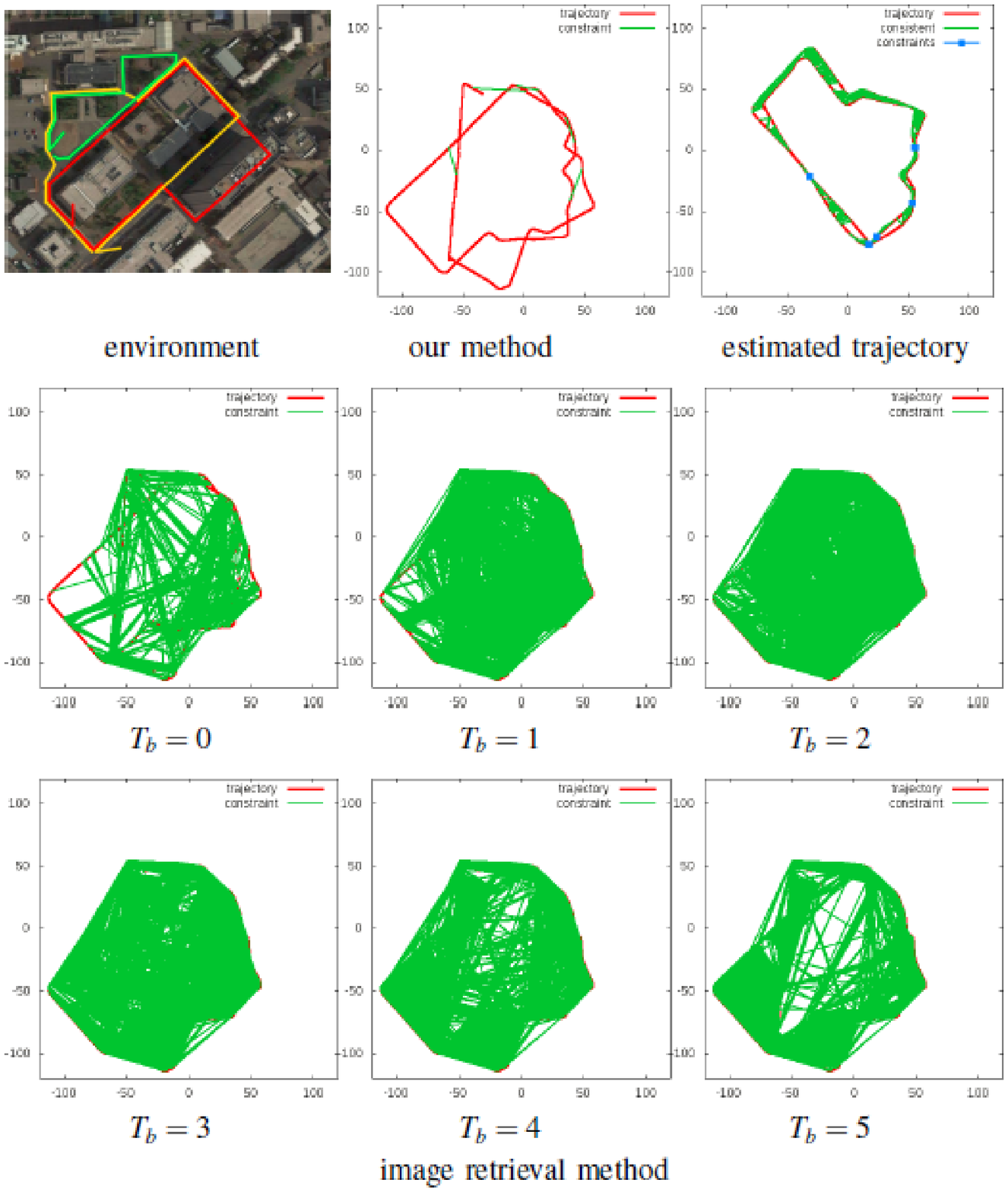}{}
    \caption{
Loop closure detection (1st row) Left: environment with robot trajectories. Red, yellow, and green lines indicate viewpoint paths on which datasets 1-3 were collected. Middle: four loops detected by our method. Right: trajectory estimated by loop closing using the detected loops as constraints. (2nd, 3rd rows) Loop closure constraints detected by using information from only the image retrieval. $T_b$ is the threshold on the dissimilarity metric (i.e., Hamming distance) between visual features (i.e., binary codes from DCNN features) employed by the image retrieval system.}\label{fig:T}
\end{figure}
}

\newcommand{\figU}{
  \begin{figure*}[t]
    \begin{center}
      \hspace*{-25mm}%
      \FIGR{8}{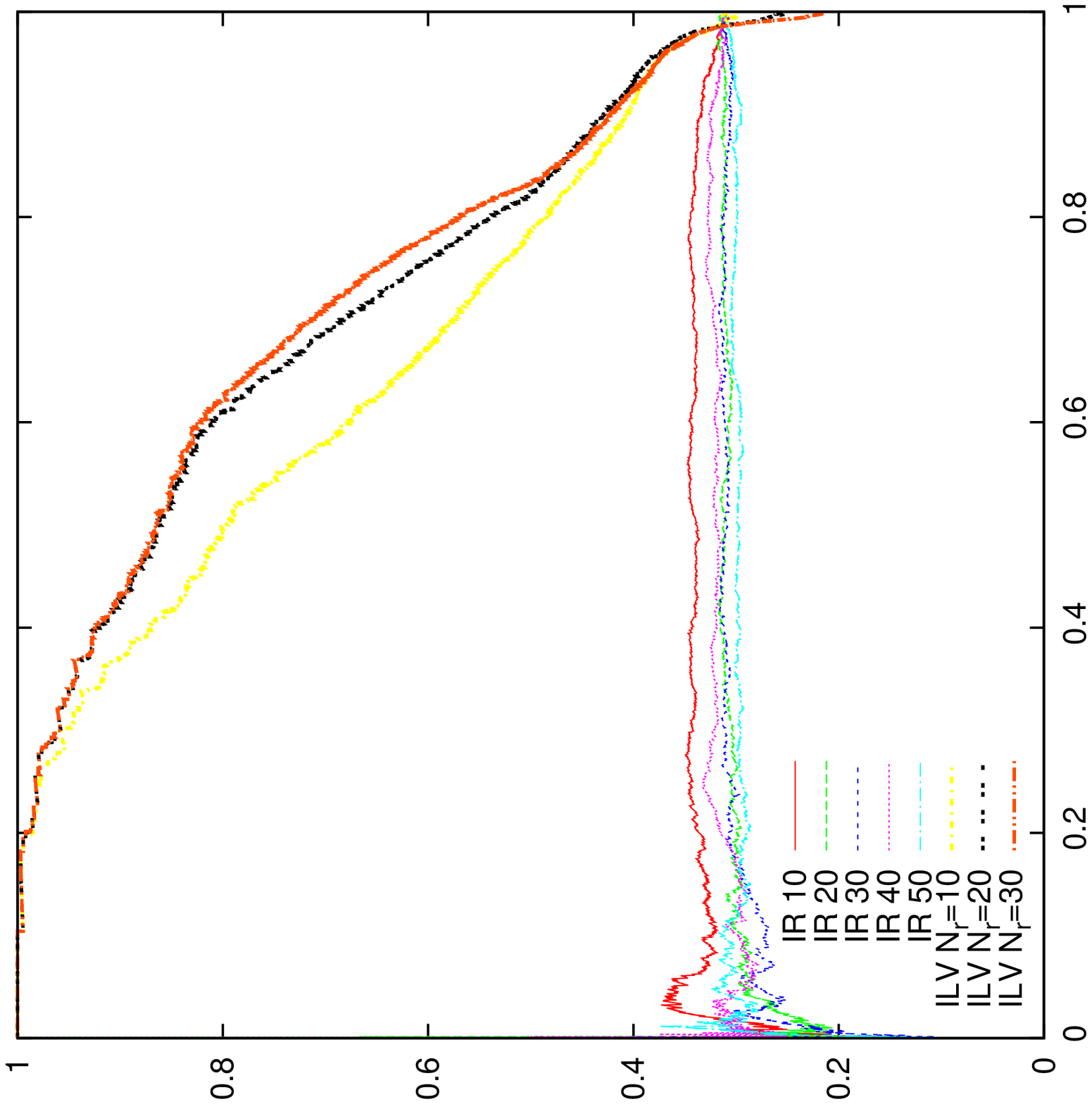}{}\hspace*{-25mm}%
\FIGR{8}{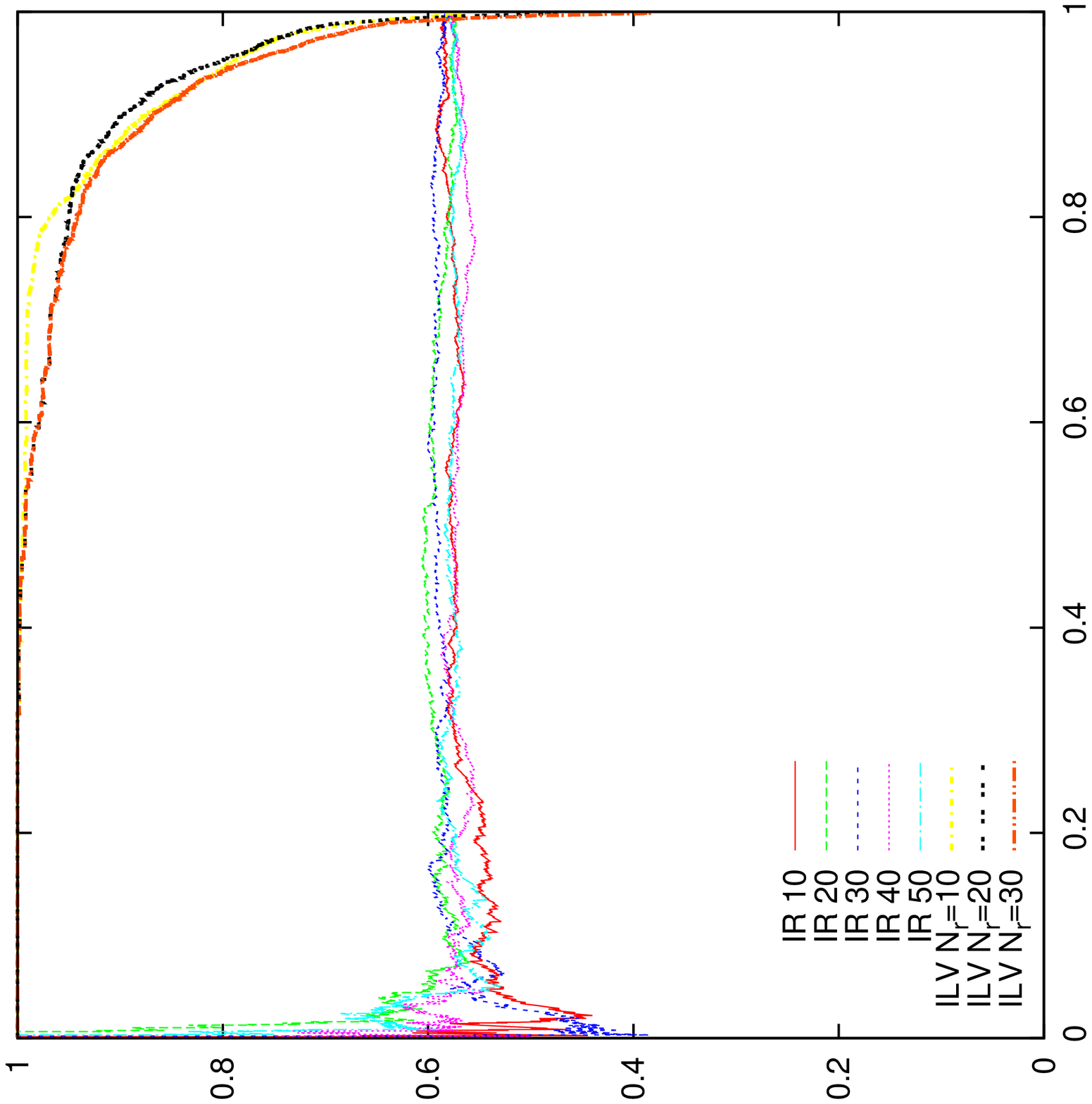}{}\hspace*{-25mm}%
\FIGR{8}{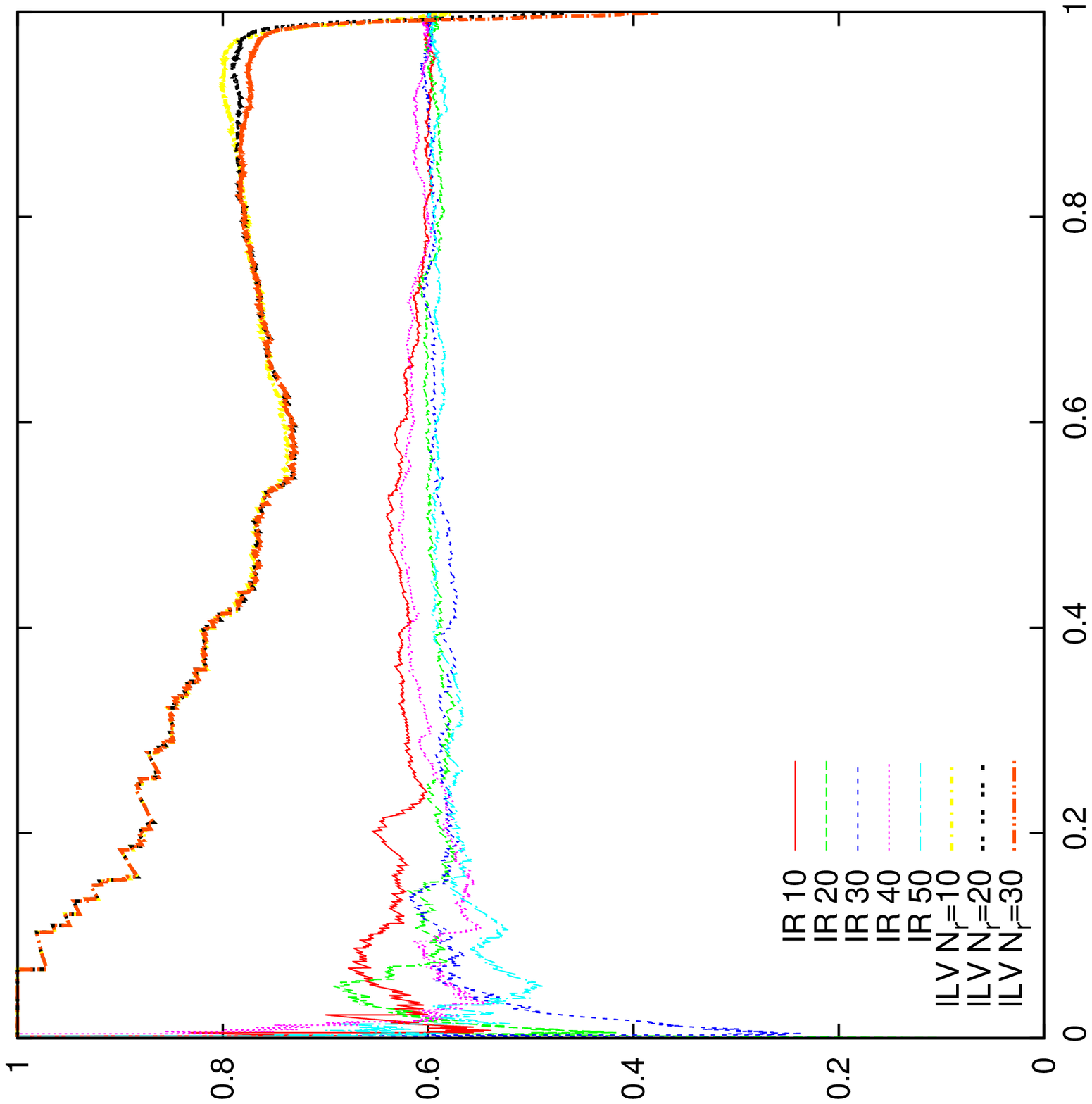}{}\hspace*{-25mm}\vspace*{-5mm}\\
\caption{Precision recall curves.
  Vertical and horizontal axes indicate precision and recall, respectively.
  IR $N_r$: image retrieval only. $N_r$ loop closure constraints randomly selected from image retrieval are used to estimate the trajectory.
  ILV $N_r$: proposed method using a different setting for $N_r$.
  Top $N_r$-ranked images from the image retrieval
  are considered as new $N_r$ VPR constraints.
}\label{fig:U}
    \end{center}
\end{figure*}
}

\newcommand{\figV}{
  \begin{figure*}[t]
    \begin{center}
\FIG{17}{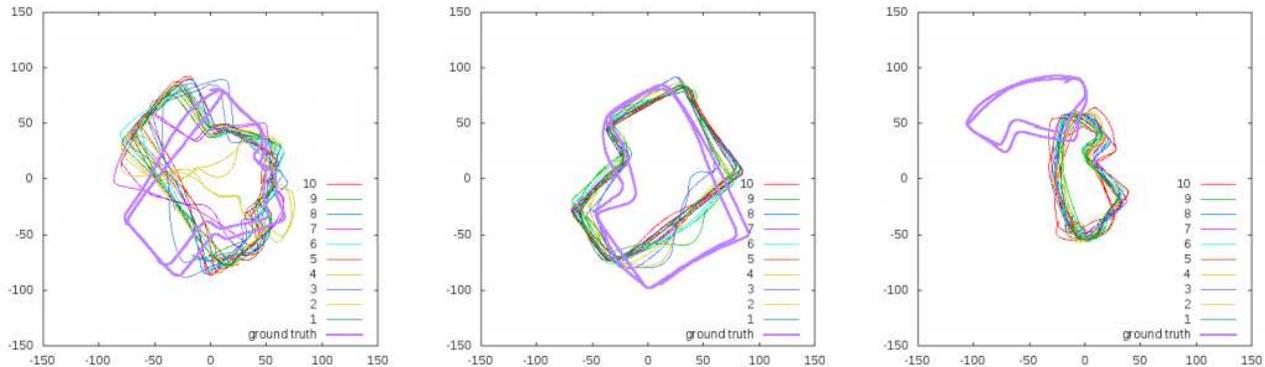}{}
\caption{Ten trajectories estimated by using
  10 hypotheses
  that received top-10 ranks at the end of navigation.
  Each colored line corresponds to the $i$-th ranked hypothesis.
  The thick line shows the ground truth trajectory.
}\label{fig:V}
\end{center}
\end{figure*}
}

\newcommand{\figW}{
  \begin{figure}[b]
    \begin{center}
\FIG{8}{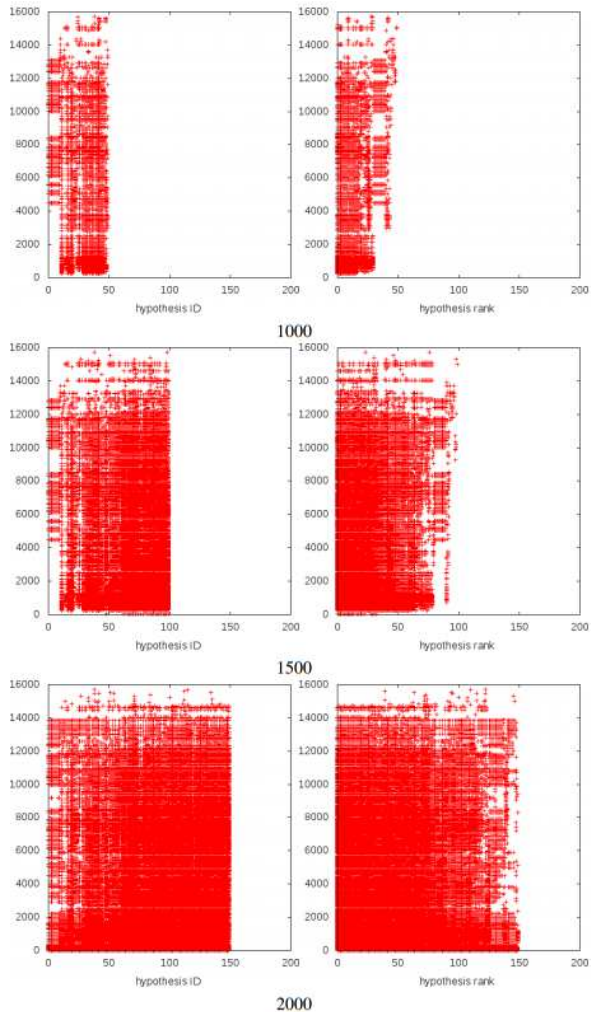}{}
\end{center}
    \caption{Consistent hypothesis-constraint pairs for different time windows
of 1000, 1500 and 2000.
Vertical axis: constraint ID. 
Horizontal axis: hypothesis ID (left) and hypothesis rank (right). 
}\label{fig:W}
\end{figure}
}

\newcommand{\figX}{
  \begin{figure}[t]
    \begin{center}
      \FIG{8}{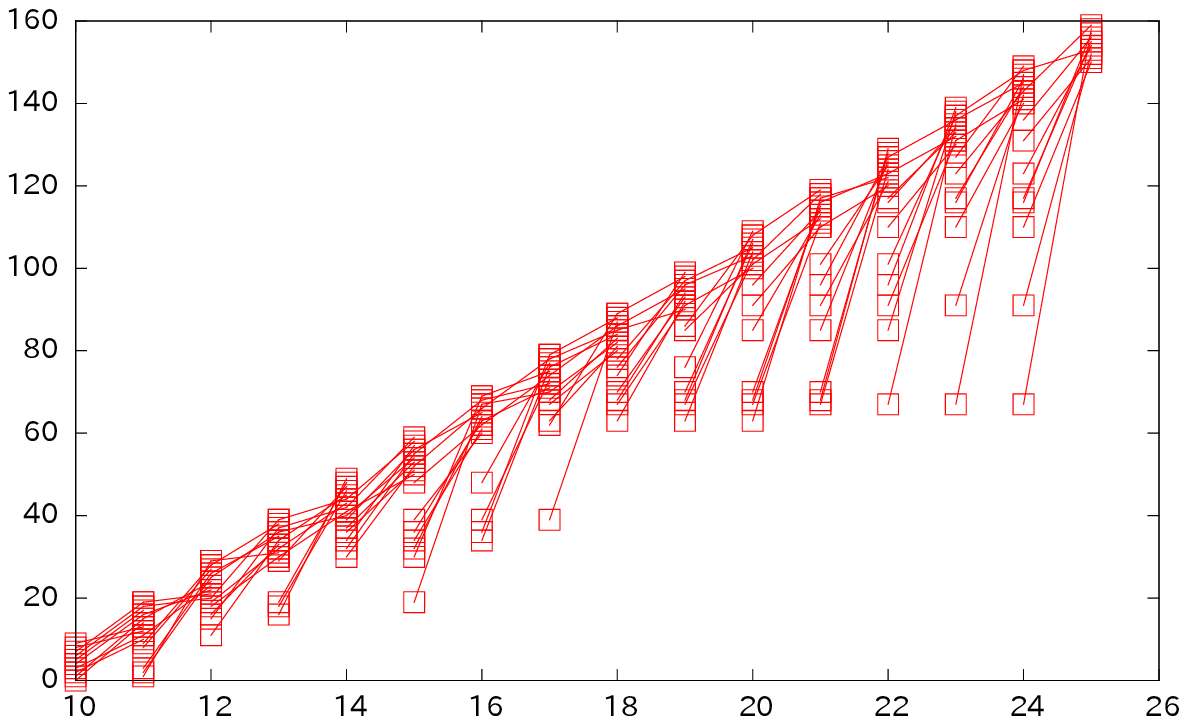}{}
      \caption{Sequential hypothesis generation.
        Vertical axis: hypothesis ID. Horizontal axis: iteration ID.
        Each line connects
        one of the top-10 ranked hypotheses
        and to the next generation hypothesis generated from it.
}\label{fig:X}
    \end{center}
    \end{figure}
  }

\newcommand{\figUa}[2]{%
\begin{minipage}[b]{4cm}
\FIGpng{3.4}{#2.png}{}{500}{385}\vspace*{-1mm}\\
\hspace*{-2mm}\FIGR{3.5}{#1_#2.eps}{}%
\end{minipage}%
\hspace*{-5mm}%
}

\title{\LARGE \bf
  Multi-Model Hypothesize-and-Verify Approach for Incremental Loop Closure Verification
} 

\maketitle

\begin{abstract}
Loop closure detection, which is the task of identifying locations revisited
by a robot in a sequence of odometry and perceptual observations,
is typically formulated as
a visual place recognition (VPR) task.
However,
even state-of-the-art VPR techniques
generate a considerable number of false positives 
as a result of confusing visual features and perceptual aliasing.
In this paper,
we propose a robust incremental framework
for loop closure detection,
termed incremental loop closure verification.
Our approach reformulates
the problem of loop closure detection 
as an instance of a 
multi-model hypothesize-and-verify framework,
in which
multiple loop closure hypotheses
are generated and verified 
in terms of the consistency
between loop closure hypotheses
and VPR constraints
at multiple viewpoints along the robot's trajectory.
Furthermore,
we consider the general incremental setting of loop closure detection,
in which 
the system must update both the set of VPR constraints
and that of loop closure hypotheses
when new constraints or hypotheses arrive during robot navigation.
Experimental results using a stereo SLAM system
and DCNN features
and visual odometry
validate effectiveness of the proposed approach.
\end{abstract}

\section{Introduction}\label{sec:intro}

Loop closure detection, 
which is the task of identifying locations revisited by
a robot in a sequence of odometry and perceptual observations, 
is a major first step to 
robotic mapping, localization and SLAM \cite{lcd2}.
Failure in loop closure detection can yield catastrophic damage in
an estimated robot trajectory, and achieving an acceptable tradeoff between
precision and recall is critical in this context. 
Loop closure detection is typically formulated as
a visual place recognition (VPR) task.
However,
even state-of-the-art VPR techniques
generate a considerable number of false positives 
as a result of confusing visual features and perceptual aliasing.

\noeditage{
  \figT
  }

In this study,
we propose a robust incremental framework
for loop closure detection,
which we call 
incremental loop closure verification.
Our approach reformulates loop closure detection as an
instance of a multi-model hypothesize-and-verify problem
in which a
set of hypotheses of robot trajectories is generated
from VPR constraints using a general pose graph SLAM \cite{thrun2005probabilistic},
and verified for
the consistency
between loop closure hypotheses
and VPR constraints
at multiple viewpoints along the robot's trajectory.
Furthermore,
we consider the general incremental setting of loop closure detection,
in which
the system must update both 
the set of VPR constraints
and that of loop closure hypotheses
when new constraints or hypotheses arrive during robot navigation.

\noeditage{
  \figR
}

The proposed approach is motivated by three independent
observations. First, we are inspired by the recent success of hypothesize-and-verify techniques (e.g., USAC \cite{raguram2013usac}).
Second, loop closure detection is essentially a multi-model
estimation problem \cite{kanazawa2004detection}, 
rather than a single model estimation considered in classical applications of the hypothesize-and-verify approach
(e.g., structure-from-motion \cite{raguram2013usac}), 
where the goal is to identify multiple
instances of models (i.e., loop closure hypotheses) and 
where the inliers to
one model 
behave as pseudo-outliers to the other models.
Finally, and
most importantly, the framework is sufficiently general and effective
for implementing various
hypothesize-and-verify
strategies that implement
various types of domain knowledge.

Although the proposed approach is general, we focus on a challenging SLAM
scenario to demonstrate the efficacy of the proposed system. Our experiments employ a stereo SLAM system that implements stereo visual odometry as in \cite{geiger2011stereoscan}, loop closure detection using
appearance-based image retrieval with DCNN features as in \cite{lcd5} 
and binary landmark as in \cite{ikeda2010visual},
and finally,
pose graph SLAM as in \cite{thrun2005probabilistic}.
Fig.\ref{fig:T} illustrates an odometry-based robot trajectory, 
as well as a trajectory corrected by loop closing, and a set of VPR
constraints selected by our method.
As can be seen, 
major errors in trajectory are accumulated as the robot navigates
and these errors are corrected given correct VPR constraints.
Fig. \ref{fig:T} also reveals 
that image retrieval-based loop closure detection
is less than perfect;
a considerable number of false positives and negatives are present.
These two types of errors, 
(i.e., accumulated errors in odometry and misrecognition in VPR of errors), 
are the main error sources that we address in this study. 
Experimental results
using our stereo SLAM system confirm that the proposed strategy, 
which includes the use of VPR constraints and loop closure hypotheses as a guide, achieves
promising results despite the fact 
that many false positive constraints and hypotheses exist.


%

The proposed approach is orthogonal to many existing approaches to loop closure detection.
In literature, most of the existing works on loop closure detection have focused on
the image retrieval step in the task, rather than the verification step
\cite{lcd2}. In fact, loop closure detection techniques are typically classified
in terms of image retrieval strategies (rather than post-verification
strategies) \cite{lcd3}. Images are typically represented by a collection of
invariant local descriptors \cite{lcd5} or a global holistic descriptor \cite{lcd6}. Loop closure detection has been employed by many SLAM
systems \cite{lcd7}. However, the above works did not focus on
the post-verification step or introduce novel insight to the hypothesize-and-verify framework.

This paper is a part of our studies on loop closure detection.
In 
\cite{arxiv15a},
we addressed the issue of guided sampling
strategies for loop closure verification,
by which the number of image matching over all the locations is minimized,
whereas the current study
assumes a constant number of image matching at each location
and focuses on the hypothesize-and-verify approach to loop closure verification.
The proposed is partially inspired by
an incremental extension of multi-modal hypothesize-and-verify approach in
\cite{icra06a},
where a different task of robot relocation was considered.
Recently, we have discussed
robust VPR algorithms \cite{iros15a}, landmark discovery \cite{icra15a}, and DCNN landmarks \cite{iros16a} in IROS15, ICRA15 and IROS16 papers. Hypothesize-and-verify strategy in loop closure detection has not been addressed in the above papers.

\section{Approach}

\subsection{Loop Closure Detection}\label{sec:3a}

For clarity of presentation, we first describe a baseline SLAM
system in which the proposed approach is built and
used as a benchmark for performance comparison in the experimental
section. 
As previously mentioned, we build the proposed system on a stereo SLAM
system in which a stereo vision sensor is employed for both 
visual odometry \cite{geiger2011stereoscan} and visual feature acquisition \cite{lcd5}.
In addition, we follow the
standard formulation of pose graph SLAM \cite{thrun2005probabilistic}. 
In pose graph SLAM,
the robot is assumed to move in an unknown environment, along a
trajectory described by a sequence of variables 
$x_{1:T}=$ $x_1$, $\cdots$, $x_T$.
While moving, it acquires sequences of odometry measurements
$u_{1:T}=$ $u_1$, $\cdots$, $u_T$
and perceptual measurements
$z_{1:T}=$ $z_1$, $\cdots$, $z_T$. 
Each odometry measurement
$u_t$ $(1\le t\le T)$
is a pairing of rotation
and translation acquired by visual odometry. Each
perceptual measurement
$z_t$
is both a set of VPR constraints $z_t^1$, $\cdots$, $z_t^{N_r}$,
and 
a pair of location IDs, $t$, $t'$.
This measurement also has a similarity score that represents 
the likelihood that the location pair belongs to the same place.
In our case, this score is obtained by the VPR task. 
More formally, we begin with an empty history of VPR constraints.
At each time $t$,
we run the VPR using the latest visual image as
a query to identify the top 
$N_r=10$
ranked retrieved images that obtain the highest
similarity scores. We then insert the
$N_r$
pairs from the query image and each of the
$N_r$
top-ranked
retrieved images as new constraints to the history.
To prevent trivial VPR constraints,
images are acquired when the interval 
$[t-\Delta t, t+\Delta t]$
is not considered a candidate for the VPR constraint ($\Delta t=200$).

For simplicity, we begin by assuming that fixed sets of
VPR constraints
$z_{1:T}$ and
$N_h$ trajectory hypotheses $m_{1:M}$ 
are a priori given. Typical hypothesize-and-verify algorithms require such a fixed set assumption \cite{kanazawa2004detection}. 
Clearly, this assumption is violated in our SLAM applications as both the VPR constraints and trajectory hypotheses must be incrementally derived as the robot navigates.
This incremental setting is addressed in Section
\ref{sec:3c} 
by relaxing
the fixed set assumption. We divide the entire measurement sequence
into constant time windows and generate
$N_g=100$
new hypotheses per window.
To generate a hypothesis, we employ pose graph SLAM that expects the following as
input: 
1) an existing trajectory hypothesis 
and 
2) a new VPR constraint selected from the history $z_{1:T}$ of loop closure constraints.
In experiments,
we set the time window size to $W=500$ frames.

The performance is evaluated in terms of quality of estimated trajectory.
As previously mentioned,
a trajectory hypothesis can be obtained
by performing the pose graph SLAM
using the loop closure hypothesis as input.
To evaluate the quality of a given trajectory hypothesis,
we first compute a set of VPR constraints
that are consistent with the hypothesis,
count the numbers of 
true positives $N_{TP}$,
false positives $N_{FP}$,
and false negatives $N_{FN}$.
We then evaluate
the precision and recall
in the form of
$N_{TP}/(N_{TP}+N_{FP})$
and $N_{TP}/(N_{TP}+N_{FN})$,
respectively.
This performance measure requires a set of ground
truth VPR constraints. 
For each query image
$i$,
we define
pairs of locations $(t,t')$
that satisfy
\begin{equation}
|| p(t,h) - p(t',h) || < T_p \label{eq:gt}
\end{equation}
as ground-truth VPR constraints.
In (\ref{eq:gt}),
$p(t,h)$ is the two-dimensional coordinate of location $t$ conditioned
on a robot trajectory hypothesis $h$,
and $T_p$ is the preset threshold of 10 m.

\subsection{VPR Constraints}\label{sec:constraints}

The image retrieval system
encodes the image to a DCNN feature representation as in \cite{lcd5}.
First,
we extract a 4,096 dimensional DCNN feature from an image.
Although a DCNN
is composed of several layers 
in each of them responses from
the previous layer are convoluted and activated by a differentiable function.
We use the sixth layer of DCNN, 
because it has proven to produce
effective features with excellent descriptive power in previous
studies \cite{lcd5}.
We then perform PCA compression to obtain 128 dimensional features.
Our strategy is supported by the recent findings
in \cite{lcd5} in which PCA compression provides excellent short codes
with 128 short vectors that generate state-of-the-art accuracy on
several recognition tasks.
In our experiments, we use DCNN
features from the image collection to train PCA models for different
settings of the output dimension of 128.
However, direct use of DCNN features for image retrieval
is computationally demanding as it
requires many-to-many comparisons
of high-dimensional DCNN features
between the query and the image collection.
To address this concern, we employ a compact 
binary encoding of images
and fast bit-count operation that enables fast image comparison (Fig. \ref{fig:R}). 
Query and
library features are encoded to $N_b=20$-bit binary codes using the compact
projection technique borrowed from \cite{tomomi2011incremental}
and then compared using the Hamming distance
to obtain a set of candidate images.
The,
L2 distance of high-dimensional DCNN features
between the query and each element of the set 
are then computed and the
top-$N_r$
elements
with the lowest L2 distance
are inserted into the constraint history $z_{1:T}$.
In this study,
the threshold $T_b$ for the Hamming distance
and the parameter $N_r$
are empirically set to 3 and 10,
respectively.
Fig.\ref{fig:R} shows several examples of input images, DCNN features and binary codes.

\subsection{Hypothesization}\label{sec:hypothesization}

At each iteration,
the system generates a set of $N_h$
hypotheses from
$N_h$ constraints
that are selected from the constraint history $z_{1:T}$.
The task of selecting $N_h$ constraints
can be formulated as
an iterative process of
selecting the $i$-th constraint
given a sequence $C^{(i-1)}$ of $(i-1)$ constraints
chosen thus far.
Therefore,
the remaining problem is
the manner by which select the $i$-th constraint.
To address this issue,
we present a simple strategy
using the trajectory hypothesis.

Our approach
begins with an empty sequence $C^{(0)}$
of constraints.
It randomly selects
the first constraint $z^{(1)}$
from the constraint history $z_{1:T}$.
We then run the pose graph SLAM
using the selected VPR constraint
to generate a trajectory hypothesis $h^{(1)}$
that is consistent with $z^{(1)}$.
Intuitively,
the next constraint $z^{(2)}$
should be {\it inconsistent}
with the trajectory hypothesis $h^{(1)}$,
because
we want to obtain
a new trajectory hypothesis $h^{(2)}$
that is dissimilar from the existing hypothesis $h^{(1)}$.
In general,
the $i$-th constraint
should be 
inconsistent with the trajectory hypothesis $h^{(i-1)}$.
To implement this idea,
we propose to select
the next constraint $z^{(i)}$
from those constraints $\{(t,t')\}$
whose distance exceeds a pre-defined threshold:
\begin{equation}
|| p(t,h^{(i-1)}) - p(t',h^{(i-1)}) || > T_p.
\end{equation}

\noeditage{
  \figQ
}

\noeditage{
\figS
}

\subsection{Incremental Extension}\label{sec:3c}

In this section, we relax the fixed set assumption given in Section \ref{sec:3a} and
consider the general incremental setting of loop closure detection,
in which 
the system must update the sets of
VPR constraints and loop closure hypotheses.
Most part of the proposed algorithm work properly with the incremental setting.

One exception is that the hypothesis list and constraint history are no longer fixed but grow linear to the time. The main space cost for a hypothesis includes 1) IDs of VPR constraints that are used for generating the hypothesis, and 
2) estimated the trajectory or a sequence of estimated robot poses in 3DOF, which is linear to time. Because the number of hypotheses is also linear to time, the total space cost grows quadratically with time.
Fortunately, in this study, the length of VPR constraints is shorter than 2000
and the number of hypotheses is smaller than 200.
As a result,
the number of evaluated hypothesis-constraint pairs is small.
Specifically,
it is less than $2000\times 200=400,000$ per iteration.
Furthermore, we observe that one can eliminate the lowest ranked hypotheses to save space and time costs, as their contribution to the overall performance is typically minimal. Time cost is nearly constant. 
Although strictly speaking, the evaluation of hypotheses using constraints grows  quadratically with time, this evaluation can be performed quickly using a look-up table 
that stores the results of image retrieval (i.e., indicating which constraint is consistent with which hypothesis). This look-up table must be updated when new hypotheses and constraints arrive, 
a process that requires a constant cost per time.

Another exception is 
with respect to the hypothesis generation procedure.
For each time window,
we iterate the hypothesis generation step for 10 times.
Then, in each hypothesis generation step, 10 hypotheses are generated from 10 top-ranked hypotheses.
This yields $N_g$$=$$10\times 10$ new hypotheses per time window.

Fig.\ref{fig:Q} and \ref{fig:S} show
examples of incremental hypothesis generation.

\section{Experiments}\label{sec:4}

\subsection{Settings}

We conducted loop closure detection experiments using a stereo SLAM system in a university campus. Our experiments employed a stereo SLAM
system that implemented the proposed strategies for loop closure detection.
The main steps involved visual odometry, loop closure detection, and post-verification. 
The first step executed stereo visual odometry in order to
reconstruct the robot trajectory. We adopted the stereo visual odometry
algorithm proposed in \cite{geiger2011stereoscan}, which has proven to be effective in recent visual odometry
applications (e.g., \cite{brubaker2013lost}). The second step applied the appearance-based
image retrieval with DCNN features,
a fast visual search using a binary landmark \cite{ikeda2010visual}
and precise image matching using DCNN features \cite{lcd5}.
This second step generates a set
of new $N_r$ VPR constraints and inserted these
into the history of VPR constraints
as mentioned in Section \ref{sec:constraints}.
The third step performed
incremental loop closure verification
to generate and verify hypotheses
in terms of the consistency
between loop closure hypotheses
and VPR constraints
from multiple viewpoints along the robot's trajectory.
This step also incorporates new hypotheses and constraints from the VPR task.

\noeditage{
\figU
}

\noeditage{
\figV
}

\noeditage{
\figW
}

\noeditage{
\figX
}

Fig. \ref{fig:T} presents robot trajectories superimposed on Google map imagery. The
ground-truth trajectories were generated using a SLAM algorithm based
on the graph optimization in \cite{thrun2005probabilistic} using manually identified ground-truth
VPR constraints as input. As indicated, major odometry errors
were collected as the robot navigated. 
We collected three sequences
along routes with travel distances of
756,
657, and
401 m, 
respectively,
using
a cart equipped with a Bumblebee stereo vision camera system, as
illustrated in Fig.\ref{fig:S}.
We defined a ground-truth loop closure constraint
as a pairing of two locations $i,j$ whose distance was less than 10 m.
Occlusion was severe in the scenes and people and vehicles were
dynamic entities occupying the scenes. We processed each path and collected
three stereo image sequences with lengths 
of 1651, 1621, and 979,
respectively.
We used images with 
a size of $640\times 480$ pixels from the
left-eye view of the stereo camera as
input for the image retrieval system.

\subsection{Results}

Fig.\ref{fig:U} shows the performance of loop closure detection in terms of estimated trajectory accuracy.
We can see that the proposed method (ILV) achieves good tradeoff between precision and recall. 
No clear correlation exists between the parameter $N_r$ and 
the performance of the proposed method. The figure also shows results for loop closure detection using only image retrieval 
(i.e., combining the binary landmark \cite{ikeda2010visual} and DCNN feature \cite{lcd5}). 
In this case, we randomly sampled $X$ ($X\in \{10,20,30,40,50\}$) VPR constraints from the history of VPR constraints and used the samples to estimate the trajectory using the pose graph SLAM. 
Fig.\ref{fig:U} reveals that these methods (i.e., loop closing with image retrieval only) do not achieve high precision performance, 
which indicates the effectiveness of the proposed post-verification framework.

Fig.\ref{fig:V} provides examples of estimated trajectories 
for 10 hypotheses that were top-ranked
at the end of the navigation. 
We can see that most of the top-ranked hypotheses are successful at estimating sufficiently accurate trajectories.

Fig.\ref{fig:W} visualizes consistency between hypotheses and constraints for three time windows of 1000, 1500, and 2000. 
This figure reveals 
that the number of hypotheses is linear to time. As expected, top-ranked hypotheses (e.g., hypotheses ranked 0-10 as shown in the right panels) are consistent with a greater number of constraints than 
are those hypotheses shown in the figure.

Fig.\ref{fig:X} shows examples of the sequential hypothesis generation.

\section{Conclusions}

The main contribution of this study is
a novel robust framework for
loop closure detection,
termed incremental loop closure verification.
Our approach reformulated
the problem of loop closure detection 
as an instance of a 
multi-model hypothesize-and-verify framework,
in which
multiple loop closure hypotheses
are generated and verified using VPR results at multiple viewpoints along the robot's trajectory.
Then,
we considered the general incremental setting of loop closure detection,
where 
the system must update the set of VPR constraints
and set of loop closure hypotheses
when new constraint or hypothesis arrives during the robot navigation.
Experimental results using a stereo SLAM system
and DCNN features
and visual odometry
validated effectiveness of the proposed approach.

\editage{
  \figT
  \figR
  \figQ
  \figS
  \figU
  \figV
  \figW
  \figX
}

\bibliographystyle{IEEEtran}
\bibliography{sii2016_lcd}

\end{document}